\documentclass[10pt,journal,compsoc]{IEEEtran}
\newif\ifpeerreview

\peerreviewfalse

\usepackage[nocompress]{cite}
\usepackage{url}
\usepackage{amsmath,amssymb,graphicx,multirow,array,ragged2e}

\usepackage{gensymb}

\usepackage[switch]{lineno}
\usepackage{caption}
\usepackage{subcaption}

\newcommand{\paperID}{XXXX}
\DeclareMathOperator*{\argmax}{argmax}

\title{Multi-Reference Image Super-Resolution: A Posterior Fusion Approach}

\author{Ke Zhao,
        Haining Tan,
        and Tsz Fung Yau
}

\begin{document}

\IEEEtitleabstractindextext{%
\begin{abstract}
\justifying
Reference-based Super-resolution (RefSR) approaches have recently been proposed to overcome the ill-posed problem of image super-resolution by providing additional information from a high-resolution image. Multi-reference super-resolution extends this approach by allowing more information to be incorporated. This paper proposes a 2-step-weighting posterior fusion approach to combine the outputs of RefSR models with multiple references. Extensive experiments on the CUFED5 dataset demonstrate that the proposed methods can be applied to various state-of-the-art RefSR models to get a consistent improvement in image quality.
\end{abstract}

}

\maketitle

\IEEEraisesectionheading{
  \section{Introduction}\label{sec:introduction}
}
%
%
%
%


Single-image super-resolution (SISR) is a computer vision task that reconstructs a high-resolution (HR) image from a low-resolution (LR) image. Typically, estimating a high-resolution image from its low-resolution counterpart is an ill-posed inverse problem \cite{kathiravan2014overview}, meaning that there are infinitely many solutions that satisfy the measurements. This underdetermined nature of the problem is particularly pronounced for images with abundant high-frequency details. To reach an optimal solution to the inverse problem with respect to certain criteria, additional regularization terms need to be specified. However, no simple regularization term can be specified to cover the characteristics of all kinds of images, thus conventional SISR algorithms are usually poorly performed.

Reference-based super-resolution (RefSR) methods explicitly exploit additional information from an external HR reference image to enhance the SISR process. Intuitively, sufficient information is encoded in a reference image that contains the same content as that on the LR image to facilitate texture restoration. However, a majority of current RefSR models can only take one reference image, limiting the amount of supplementary information to incorporate into the super-resolution process.

To achieve multi-reference-based super-resolution (MRefSR), two approaches are possible: it can either be that multiple reference images are used as the initial inputs to the model, or that the multiple outputs of SRefSR using different reference images are fused to combine the information. We observed that nowadays most RefSR models aim to achieve better content alignment, both spatially and semantically, between the input image and one reference image. Because it is intuitively hard for additional reference images to contribute to the alignment process, we claim that the posterior fusion of multiple SRefSR outputs would be a more natural way to combine the relevant information from each reference image. Following this idea, we proposed a two-step-weighting fusion scheme that can be incorporated into a variety of existing SRefSR models to achieve MRefSR and better-quality final SR images. Also, our proposed method has a low computational cost and allows for parallel computation for the super-resolution process with multiple reference images.

The proposed posterior fusion method can be applied to a wide range of applications. For example, in video game graphic rendering, HR patches for each object in the scene are readily available as the texture to be mapped, and they can be used as reference images to perform MRefSR. This approach would be applied to any video game and would save huge computing resources compared to the state-of-the-art NVIDIA DLSS, which trains separate SR neural networks for each video game.

\section{Related Work}
\subsection{Single Image Super-Resolution}
With the popularity of Convolutional Neural Networks (CNN), learning-based approaches demonstrate significantly better performances given an appropriate training set. Early-stage CNN-based SISR models like SRCNN \cite{SRCNN} choose pixel-level reconstruction errors such as MSE and MAE between the recovered HR image and ground truth as loss functions to optimize. Furthermore, significant improvements can be made by optimizing the standard CNN architecture. For instance, the approach EDSR \cite{Lim_2017_CVPR_Workshops} proposed by Lim et al. 
applied the residual network architecture to the SR task and achieved superior results. While these algorithms tend to maximize the peak signal-to-noise ratio (PSNR), they often result in smooth reconstruction lacking high-frequency details and are perceptually unsatisfying. To state the problem of SISR in another way, downsampling an HR image is an irreversible compression process during which much high-frequency information is lost. Instead of trying to recover the lost information from nowhere, Ledig et al. \cite{ledig2017photo} adopt Generative Adversarial Networks (GAN) and proposed SRGAN that generates "fake" texture details that are visually realistic. While these results are perceptually satisfying, texture details in these images are hallucinations and are often different from those in the ground-truth images, resulting in PSNR degradation. This deficiency makes the methods like SRGAN unsuited for fidelity-sensitive applications like medical imaging. Additionally, pure GAN-based SISR approaches fail to produce satisfying results on test images with complicated components, often resulting in distorted color lumps.

\subsection{Single-Reference-based Super-Resolution}
Zheng et al. \cite{Zheng_2018_ECCV} proposed an end-to-end approach, named CrossNet, based on fully convolutional neural networks that can perform spatial alignment between the reference features and the LR features. One issue of this model is that regions of the reference image that are irrelevant to the input will degrade the performance. Motivated by this, Shim et al. \cite{Shim_2020_CVPR} proposed a robust RefSR model that is aware of the relevancy of the reference image, leading to a more robust result that outperforms SRGAN in terms of generating visually stratifying SR images while also achieving high PSNR. More recently, Zhang et al. \cite{zhang2022self} proposed to use dual zoomed observations (from a telephoto) as references and apply self-supervised techniques to that. This is inspired by multiple cameras in modern smartphones that are able to collect dual-zoomed observations at the same time. While this model performs well for scenes with repeated structural texture, it gives highly distorted outputs for common scenarios.

\subsection{Multi-Reference-based Super-Resolution}
MRefSR has recently been proposed to extend the idea of RefSR, and previous works in this field majorly focus on designing novel neural network models such that they take multiple reference images as the initial inputs. Yan et al. \cite{yan2020towards} proposed a content-independent MRefSR model that builds up a universal reference pool before doing predictions. Given an input low-resolution image alone, this model finds reference images whose textures are similar to each segmentation of the input from its pool to help the reconstruction process. A hierarchical attention-based sampling approach is proposed by Pesavento et al. \cite{pesavento_volino_hilton_2021} to combine the features of multiple reference images. While these models provide a promising direction, the performance improvement compared to SRefSR is still limited.

\begin{figure*}[!t]
\centering
\includegraphics[width=1.8\columnwidth]{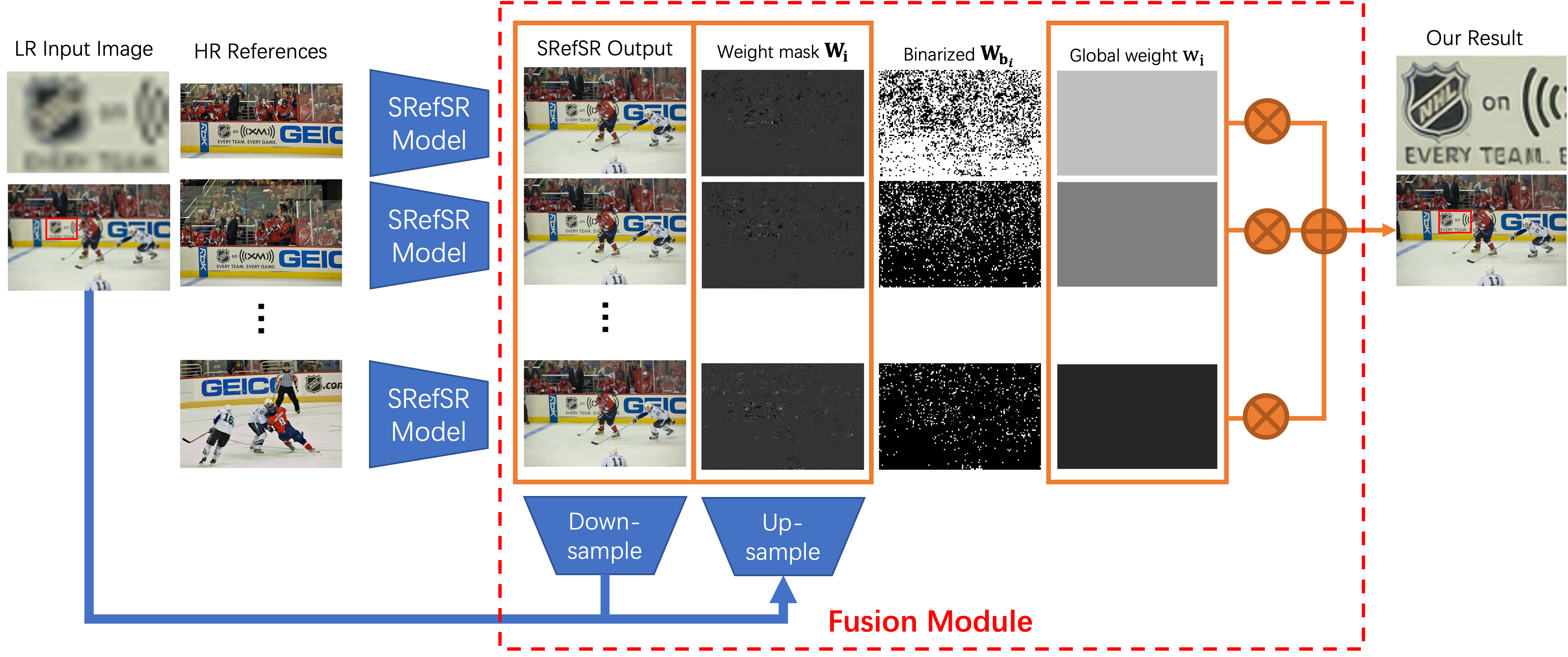}
\caption{Our proposed multi-image RefSR pipeline}
\label{fig:method_overal}
\end{figure*}

\section{Proposed Method}
The overview of our proposed multi-image RefSR pipeline is shown in Fig. \ref{fig:method_overal}. This proposed method consists of two major parts: 1) An image alignment and texture extraction module, which can be any existing single-image RefSR model, and 2) A image fusion module. The first module takes an LR input image and an HR reference image and tries to match the HR reference spatially and semantically to the LR input and then uses the corresponding HR texture in the reference image to get an HR version of the input image. By feeding one input LR image and multiple reference images to this module, multiple SR output images are obtained. The second module fuses these output images into a single SR image by combining the best region of each SR output, with the objective to improve image quality. The fusion module consists of two steps, namely adaptive weight masking and globally reference-quality-based weighted averaging.

\subsection{Naive Fusion}
Before introducing our proposed fusion method, let's first consider the simplest fusion scheme, which is just averaging the intensity of pixels of each single-reference SR output. That is, for every pixel position $p$ in the SR image $\mathbf{I}_i$ generated by the RefSR module, the corresponding pixel value of the fused image $\hat{\mathbf{I}}$ is given by:
\begin{align}
    \hat{\mathbf{I}}(p) = \frac{1}{N} \sum_{i=1}^{N} \mathbf{I}_i(p)
\end{align}
where $N$ is the total number of reference images and $\mathbf{I}(p)$ is the intensity of the pixel at position $p$. The index $i$ of SR image $\mathbf{I}_i$ is arranged such that the lowest indexed image $\mathbf{I}_1$ corresponds to the RefSR result with the most relevant reference image. 

By relevant we mean the content of the reference image is similar to that of the input image. Take the CUFED5 dataset that is adopted for evaluation by this paper as an example, the leftmost image in Fig. \ref{fig:cufed5} is the ground truth HR version of the LR input image, while the other images are the reference images whose relevance to $\mathbf{I}_{GT}$ decreases from left to right. 

It is intuitive that $\mathbf{I}_1$ would have the best quality and $\mathbf{I}_N$ would have the worst. As most of the previous single-reference SR works take the $\mathbf{I}_1$ as their final results, averaging $\mathbf{I}_1$ with the rest $\mathbf{I}_i$ will degrade the quality of the final fused image $\hat{\mathbf{I}}$. These claims are supported by our experimental results.

\subsection{Adaptive Weight Masking}
Instead of naively averaging the single-reference SR image $\mathbf{I}_i$ as a whole, a more desirable fusion method would combine the best of each SR image. As illustrated in Fig. \ref{fig:method_overal}, even though $\mathbf{I}_1$ (the top one in the middle column) has the best overall quality, the $\mathbf{I}_3$ has sharper character reconstruction on the selected region. To achieve this, an adaptive weight mask $\mathbf{W}_i$, whose dimension is the same as SR image $\mathbf{I}_i$, is computed. It gives higher weights to pixels in regions with better-quality reconstruction and each pixel value of the fused image $\hat{\mathbf{I}}$ is given by:
\begin{align}
    \hat{\mathbf{I}}(p) = \frac{1}{\sum_{i=1}^N \mathbf{W}_i(p)} \sum_{i=1}^{N} \mathbf{I}_i(p) \mathbf{W}_i(p)
\end{align}
Ideally, $\mathbf{W}_i(p)$ should measure how close a pixel is to the ground-truth image, but this information is not available in real RefSR scenarios. What is readily available is the LR input image $\mathbf{I}_{input}$ itself.

To compute $\mathbf{W}_i(p)$, $\mathbf{I}_i$ is first downsampled to have the same dimension as the input image, denoted by $\mathcal{D}(\mathbf{I}_i)$. The idea is that, for an ideal reconstruction of the input image, the downsampled version of it would be exactly the same as the input image. Therefore the difference between the pixel intensities of $\mathcal{D}(\mathbf{I}_i)$ and $\mathbf{I}_{input}$ would be a good proximity to the difference between $\mathbf{I}_i$ and $\mathbf{I}_{GT}$. Specifically, $\mathbf{W}_i$ is given by:
\begin{align}
    \mathbf{W}_i = \mathcal{U}(\exp\left(-\beta (\mathcal{D}(\mathbf{I}_i) - \mathbf{I}_{input})^2\right))
\end{align}
where $\mathcal{U}$ denotes bicubic upsampling and $exp$ is element-wise exponential. $\beta$ is a parameter controlling how much the discrepancy in pixel intensities is penalized.

\subsection{Global Reference-Quality-based Weight} \label{sec:global_weight}
While the Adaptive Weight Masking scheme works reasonably well in cases for severely distorted regions in $\mathbf{I}_i$, it is insensitive to small distortions in $\mathbf{I}_i$, which is crucial in fine detail recovery. To see why this is happening, consider the most naive super-resolution result of $\mathbf{I}_{iput}$, that is, its bicubic interpolated upsampled version. Despite having the worst SR quality, this image would be yet another optimal solution with maximized $\mathbf{W}_i$ since its bicubic downsampled version would be exactly the same as the $\mathbf{I}_{iput}$.

\begin{figure}[ht]
\centering
\includegraphics[width=0.9\columnwidth]{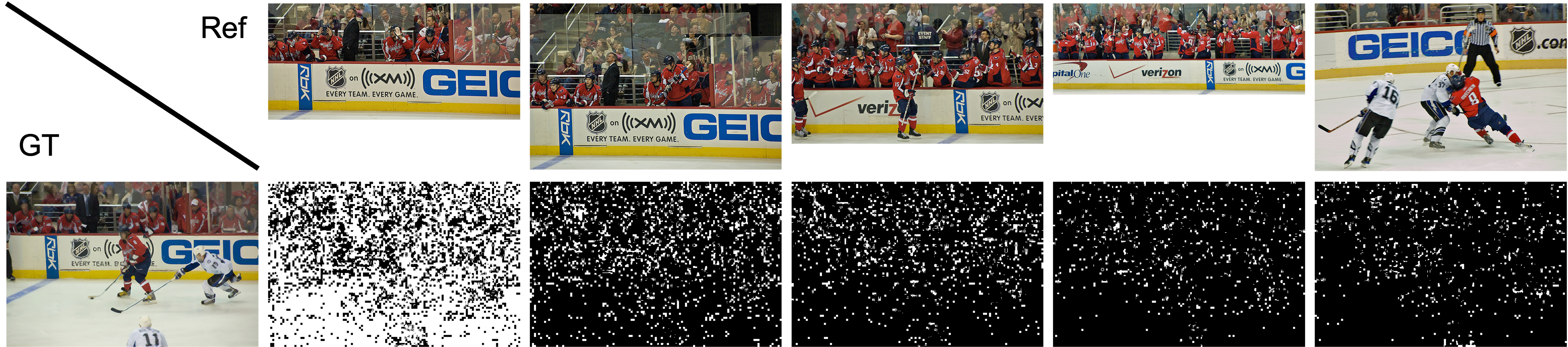}
\caption{Binary weight mask computed from RefSR results}
\label{fig:binary_mask}
\end{figure}

To overcome this problem, an measurement to encourage fidelity is needed. One might consider Natural Image Prior \cite{kim2010single} as a way to force fine details. However, we have observed the deficiencies in the RefSR could be both blurring lumps and noise high-frequency mosaic, making Natural Image Prior noneffective. Instead, we took an indirect measurement of the fidelity of $\mathbf{I}_i$. As shown in Fig. \ref{fig:binary_mask}, the black-and-white figure is the binary weight mask computed by finding the maximum value of $\mathbf{W}_i(p)$ for each pixel across all RefSR results $\mathbf{I}_i$, so the pixel intensity each the binary weight mask is given by:
\begin{align}
    \mathbf{W}_{b_{i}}(p) =
    \begin{cases}
        1 &, i = \argmax\limits_{i}{\mathbf{W}_i(p)} \\
        0 &, \text{otherwise}
    \end{cases}
\end{align}
It can be observed from Fig. \ref{fig:binary_mask} that the total area of the white region in the binary weight mask figure can be an indicator of how relevant the underlying reference image is to the input image. Intuitively, the better the underlying reference, the better the texture quality in the SR results. Therefore, the sum of the binary weight mask is adopted as a global weight for $\mathbf{I}_i$ in the fusion process. This global weight is computed by:
\begin{align}
    w_i = \exp{\left(\beta_g \sum_{p} \mathbf{W}_{b_{i}}(p)\right)}
\end{align}
And the fused image is given by:
\begin{align}
    \mathbf{I}_{fused} = \frac{1}{\sum_i w_i} \sum_i w_i {\hat{\mathbf{I}}}_i
\end{align}
where $\hat{\mathbf{I}}_i$ is the $i^{th}$ SR image after applied Adaptive Weight Masking, and $\beta_g$ is a parameter controlling how much priority is given to the $\hat{\mathbf{I}}_i$ with the best underlying referencing image.

\section{Experimental Results}
\subsection{Dataset and Evaluation Metrics}

\begin{figure}[ht]
\centering
\includegraphics[width=0.9\columnwidth]{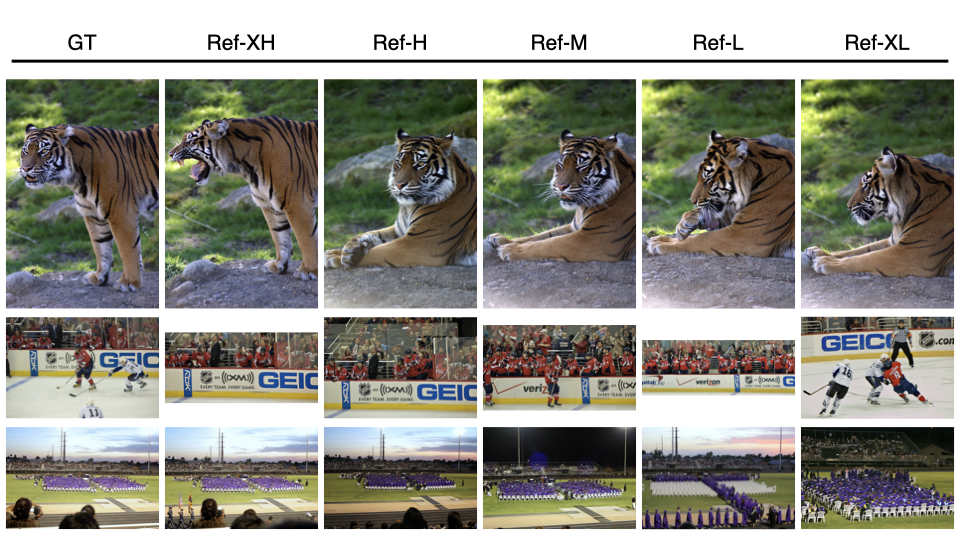}
\caption{Sample images from the test set of CUFED5}
\label{fig:cufed5}
\end{figure}

\noindent \textbf{Testing Dataset.} The test set of CUFED5 \cite{Zhang_2019_CVPR} is used as the testing dataset because it provides multiple reference images for each low-resolution image. (Fig. \ref{fig:cufed5}). The CUFED5 test set has 126 HR input images and each has 5 HR reference images with different similarity levels. The input LR images for the evaluation below are constructed by 4x bicubic downsampling from the ground-truth HR images.

\noindent \textbf{Evaluation Metrics.} The quantitative experiments adopt PSNR and SSIM (structural similarity) \cite{wang2004image} on the Y channel of the YCrCb space as evaluation metrics. In case that different RefSR modules generate images with different dimensions, the cropping/padding/interpolation scheme in the fusion module is chosen to be the same as that in the RefSR modules so that the evaluation results are consistent.

\begin{table*}[ht]
    \centering
    \begin{tabular}{m{3.3cm} m{1.2cm} m{1.2cm} m{1.2cm} m{3.3cm} m{1.2cm} m{1.2cm} m{1.2cm}}
    Input Image & HR & C$^2$-Matching & AMSA & Input Image & HR & C$^2$-Matching & AMSA\\
    \hline
    Reference Image & ESRGAN & Ours (C$^2$-Matching) & Ours (AMSA) & Reference Image & ESRGAN & Ours (C$^2$-Matching) & Ours (AMSA)\\
    
    \includegraphics[height=1.5cm]{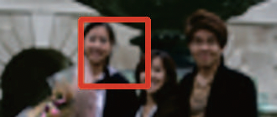} &
    \includegraphics[height=1.5cm]{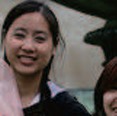} &
    \includegraphics[height=1.5cm]{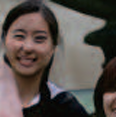} &
    \includegraphics[height=1.5cm]{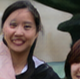} & 
    \includegraphics[height=1.5cm]{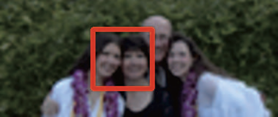} &
    \includegraphics[height=1.5cm]{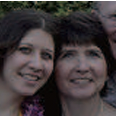} &
    \includegraphics[height=1.5cm]{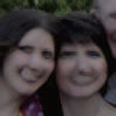} &
    \includegraphics[height=1.5cm]{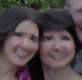} \\
    \includegraphics[height=1.5cm]{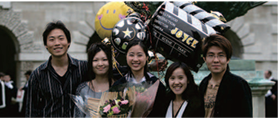} &
    \includegraphics[height=1.5cm]{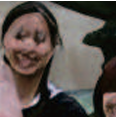} &
    \includegraphics[height=1.5cm]{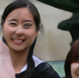} &
    \includegraphics[height=1.5cm]{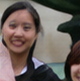} &
    \includegraphics[height=1.5cm]{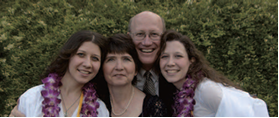} & 
    \includegraphics[height=1.5cm]{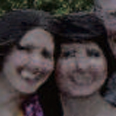} &
    \includegraphics[height=1.5cm]{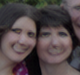} &
    \includegraphics[height=1.5cm]{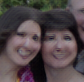} \\
    \includegraphics[height=1.5cm]{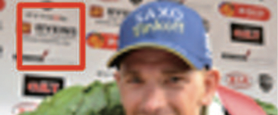} &
    \includegraphics[height=1.5cm]{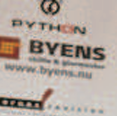} &
    \includegraphics[height=1.5cm]{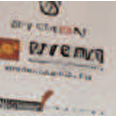} &
    \includegraphics[height=1.5cm]{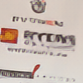} & 
    \includegraphics[height=1.5cm]{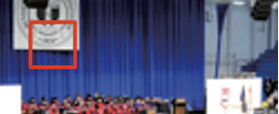} &
    \includegraphics[height=1.5cm]{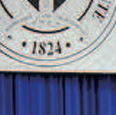} &
    \includegraphics[height=1.5cm]{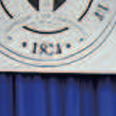} &
    \includegraphics[height=1.5cm]{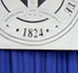} \\
    \includegraphics[height=1.5cm]{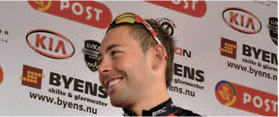} &
    \includegraphics[height=1.5cm]{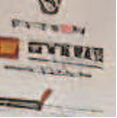} &
    \includegraphics[height=1.5cm]{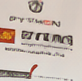} &
    \includegraphics[height=1.5cm]{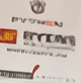} &
    \includegraphics[height=1.5cm]{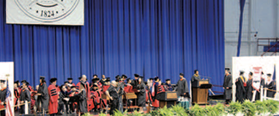} & 
    \includegraphics[height=1.5cm]{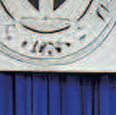} &
    \includegraphics[height=1.5cm]{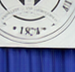} &
    \includegraphics[height=1.5cm]{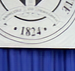} \\
    \includegraphics[height=1.5cm]{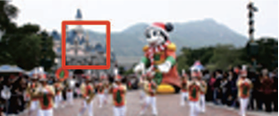} &
    \includegraphics[height=1.5cm]{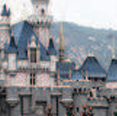} &
    \includegraphics[height=1.5cm]{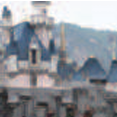} &
    \includegraphics[height=1.5cm]{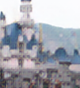} & 
    \includegraphics[height=1.5cm]{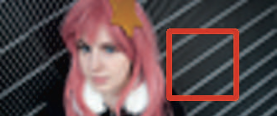} &
    \includegraphics[height=1.5cm]{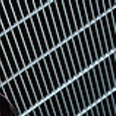} &
    \includegraphics[height=1.5cm]{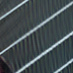} &
    \includegraphics[height=1.5cm]{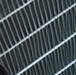} \\
    \includegraphics[height=1.5cm]{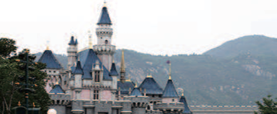} &
    \includegraphics[height=1.5cm]{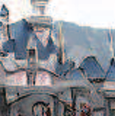} &
    \includegraphics[height=1.5cm]{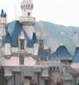} &
    \includegraphics[height=1.5cm]{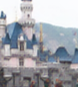} &
    \includegraphics[height=1.5cm]{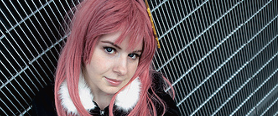} & 
    \includegraphics[height=1.5cm]{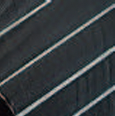} &
    \includegraphics[height=1.5cm]{QuaImages/6_c2_m.png} &
    \includegraphics[height=1.5cm]{QuaImages/6_am_m.png} \\
    \end{tabular}
    \caption{Qualitative results}
    \label{tab:quali}
\end{table*}

\subsection{Qualitative Comparisons}
Table \ref{tab:quali} summarizes the qualitative comparison with the state-of-the-arts. We applied our method on $C^2$-Matching \cite{jiang2021robust} and AMSA \cite{xia2022coarse} and compared the results with the original ones. Also, the result of ESRGAN \cite{wang2018esrgan} is included as a representation of SISR results. It can be observed that our method does a certain level of visual quality improvement upon the original work, especially in the case of $C^2$-Matching, where a denoising bonus is applied upon super-resolution.

\subsection{Quantitative Evaluation}

\begin{table}[ht]
\renewcommand{\arraystretch}{1.3}
\caption{Evaluation of the models on PSNR\_Y and SSIM}
\centering
\begin{tabular}{c||c|c}
\hline
 & Model & PSNR/SSIM\\
\hline\hline
\multirow{7}{4em}{SISR} & SRCNN & 25.33/0.745\\
& EDSR & 25.93/0.777\\
& RCAN & 26.06/0.769\\
& SRGAN & 24.40/0.702\\
& ENet & 24.24/0.695\\
& ESRGAN & 21.90/0.633\\
& RankSRGAN & 22.31/0.635\\
\hline\hline
\multirow{7}{4em}{Ref SR} & CrossNet \cite{Zheng_2018_ECCV}& 25.48/0.764\\
& SRNTT & 25.61/0.764\\
& TTSR & 25.53/0.765\\
& SSEN & 25.35/0.742\\
& E2ENT$^2$ & 24.01/0.705\\
& CIMR & 26.16/0.781\\
& C$^2$-Matching & 27.16/0.805 \\
& AMSA & 27.31/0.809 \\
\hline\hline
& Ours (with C$^2$-Matching) & \textbf{27.29/0.806} \\ 
& Ours (with AMSA) & \textbf{27.56/0.825} \\ 
\hline
\end{tabular}
\label{tab:quanti}
\end{table}

\noindent \textbf{Overall Comparisons.} Table \ref{tab:quanti} shows the quantitative comparison of the performance of a variety of super-resolution methods. We applied the proposed method to $C^2$-Matching\cite{jiang2021robust} and AMSA\cite{xia2022coarse}, and compared and results with the original works. We also include the results of representative models of SISR and RefSR, which are evaluated on the same dataset as outs. For SISR methods, we include SRCNN\cite{SRCNN}, EDSR\cite{Lim_2017_CVPR_Workshops}, RCAN\cite{zhang2018image}, SRGAN\cite{ledig2017photo}, ENet\cite{paszke2016enet}, ESRGAN\cite{wang2018esrgan} and RankSRGAN\cite{Zhang_2019_ICCV}. For RefSR methods, we include CrossNet\cite{Zheng_2018_ECCV}, SRNTT\cite{Zhang_2019_CVPR}, TTSR\cite{yang2020learning}, SSEN\cite{jiang2021robust}, E2ENT$^2$\cite{xie2020feature} and CIMR\cite{yan2020towards}.

We can see that our method outperforms all SISR and Ref SR models. In particular, $C^2$-Matching and AMSA have shown improvement after integrating with the pipeline.

\begin{figure}[ht]
    \centering
    \begin{subfigure}[b]{0.31\columnwidth}
    \centering
    \includegraphics[width=\textwidth]{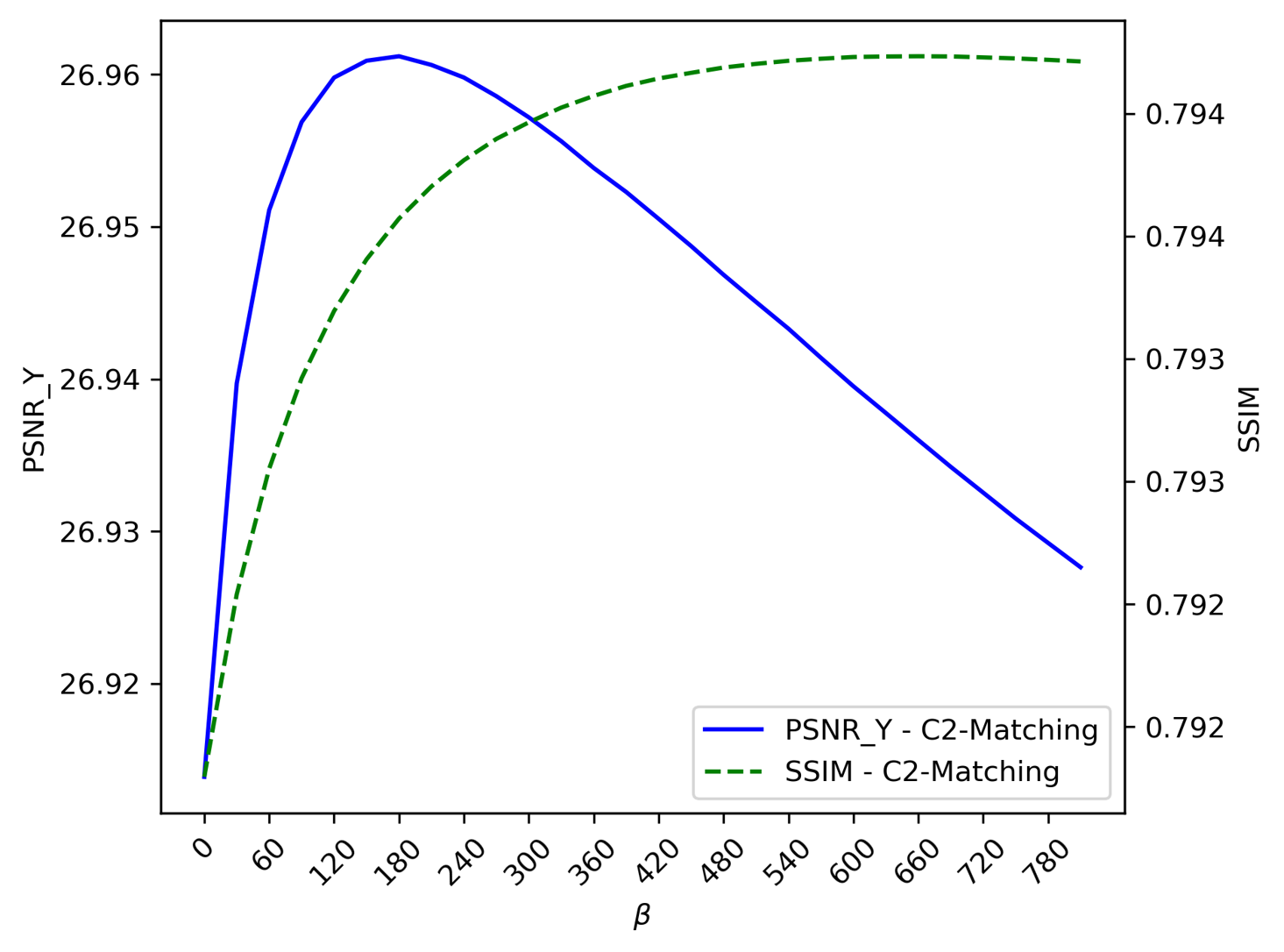}
    \caption{}
    \end{subfigure}
    \begin{subfigure}[b]{0.31\columnwidth}
    \centering
    \includegraphics[width=\textwidth]{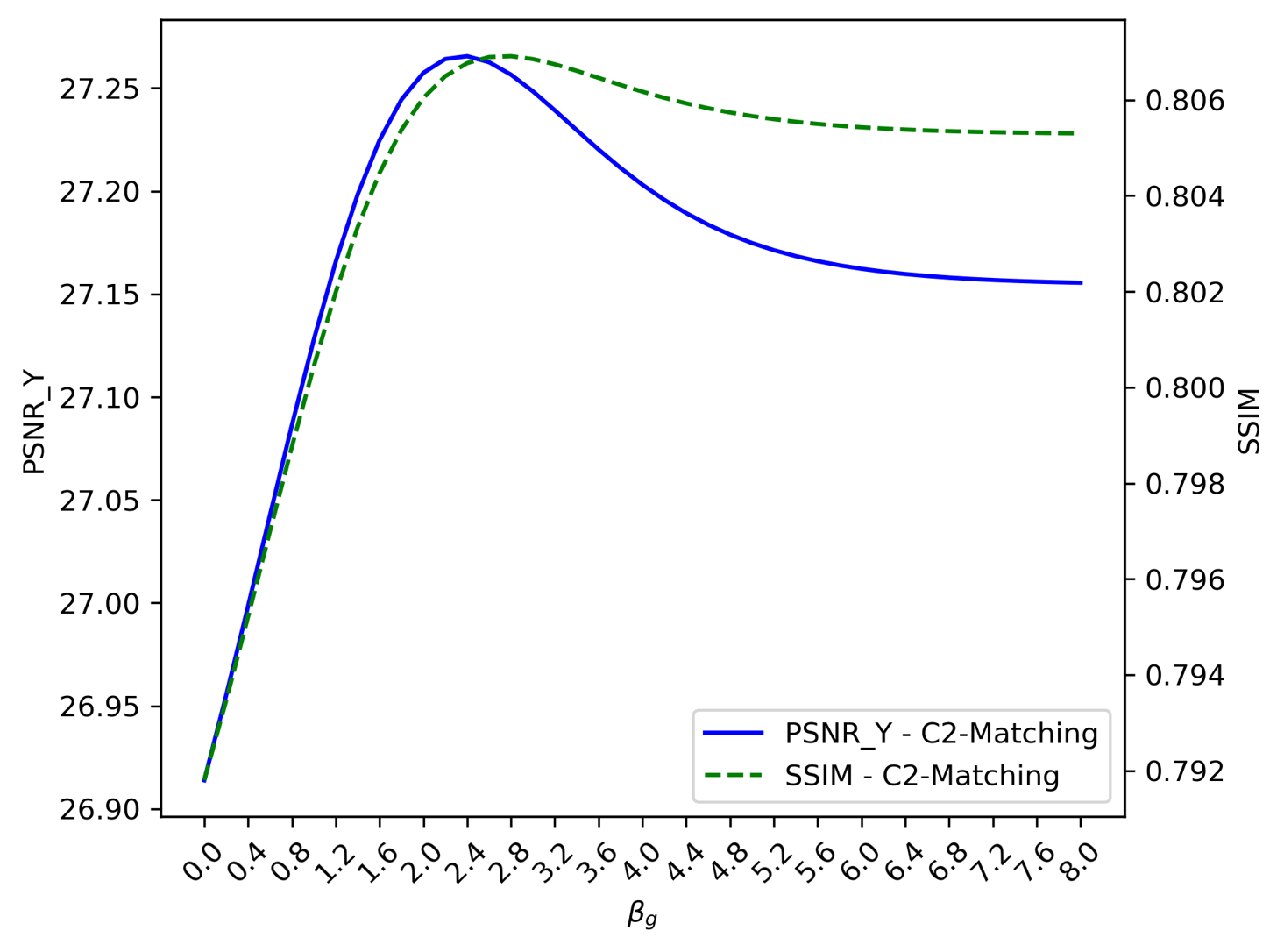}
    \caption{}
    \end{subfigure}
    \begin{subfigure}[b]{0.31\columnwidth}
    \centering
    \includegraphics[width=\textwidth]{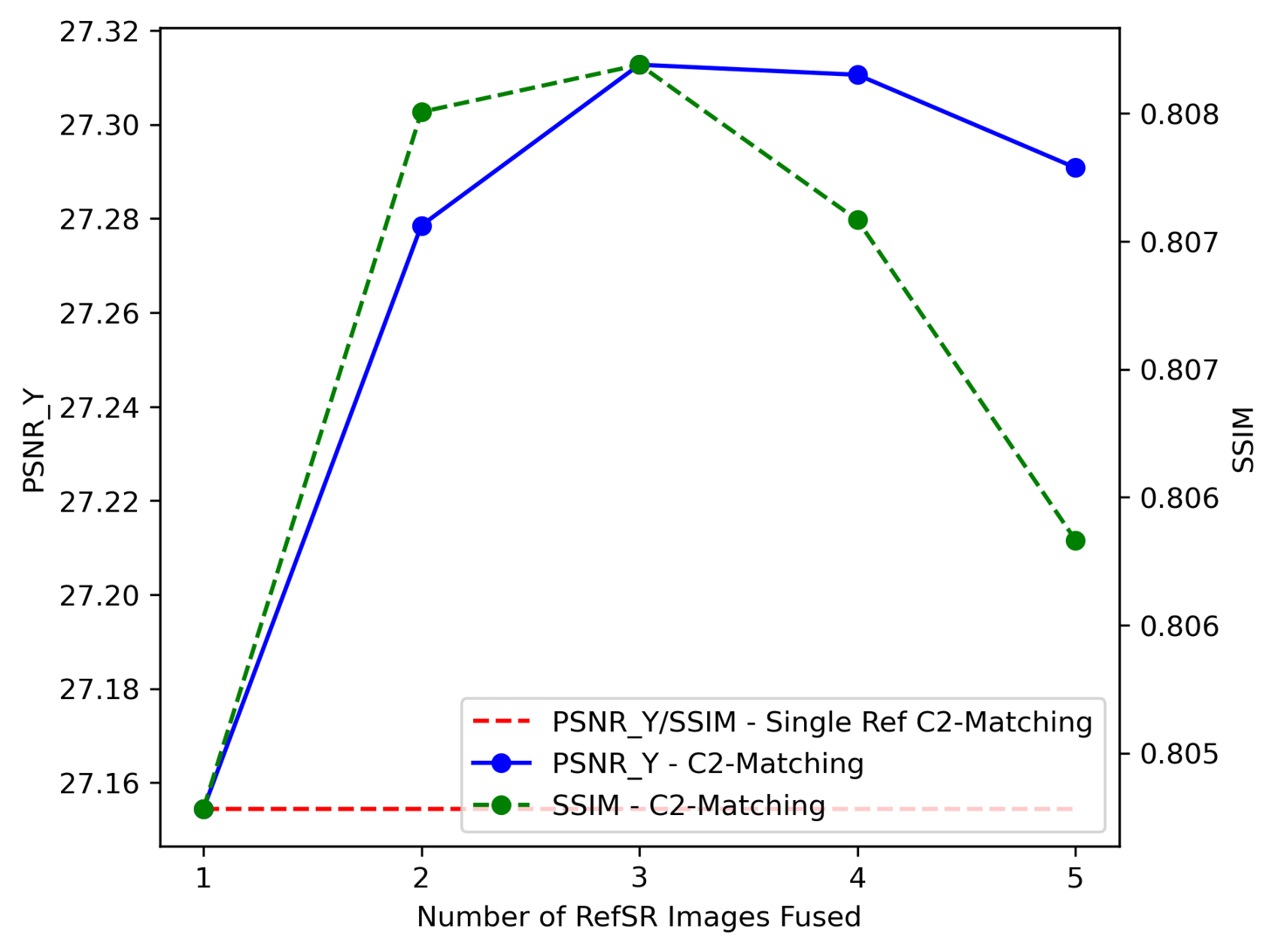}
    \caption{}
    \end{subfigure}
    
    \begin{subfigure}[b]{0.31\columnwidth}
    \centering
    \includegraphics[width=\textwidth]{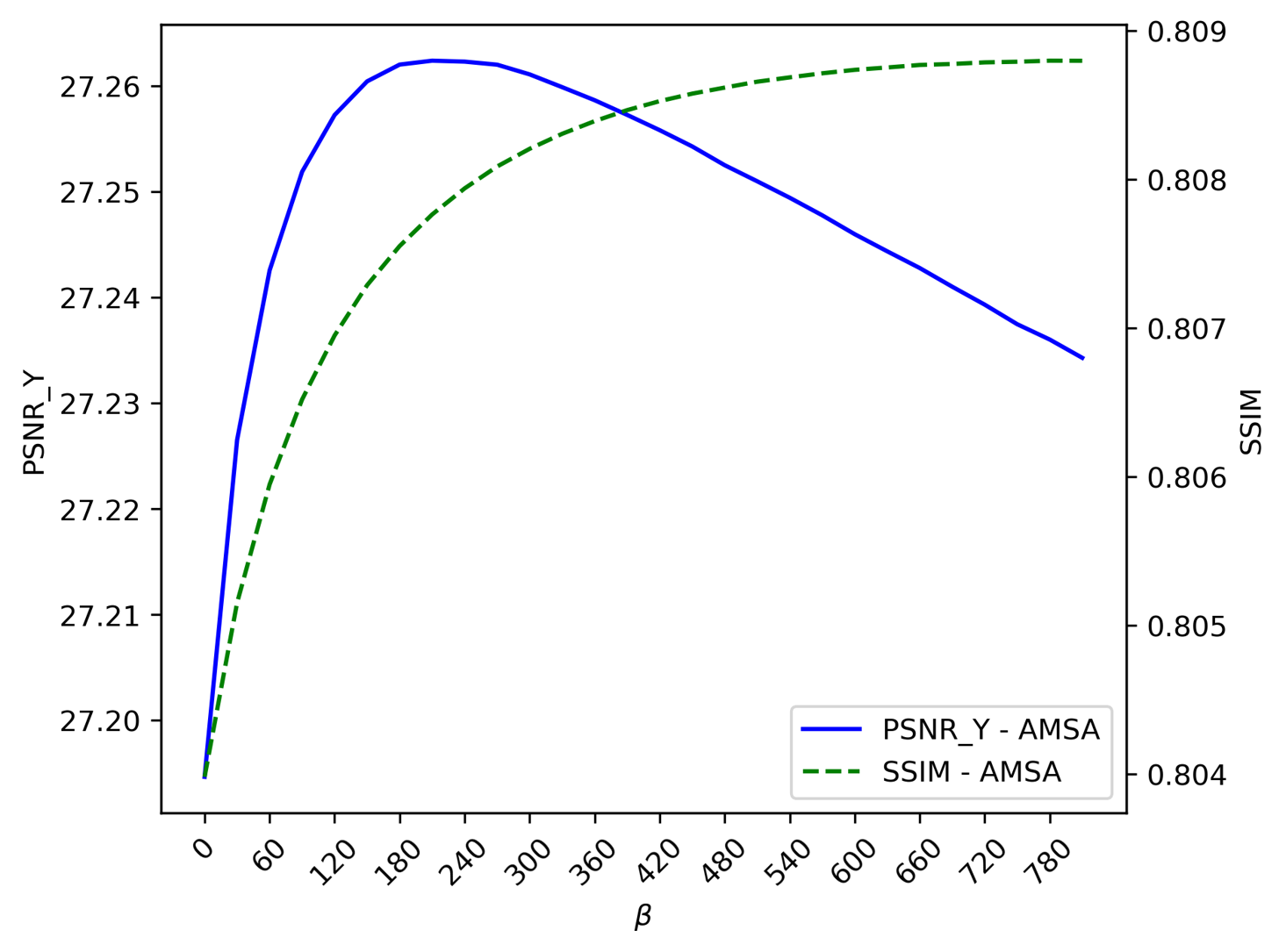}
    \caption{}
    \end{subfigure}
    \begin{subfigure}[b]{0.31\columnwidth}
    \centering
    \includegraphics[width=\textwidth]{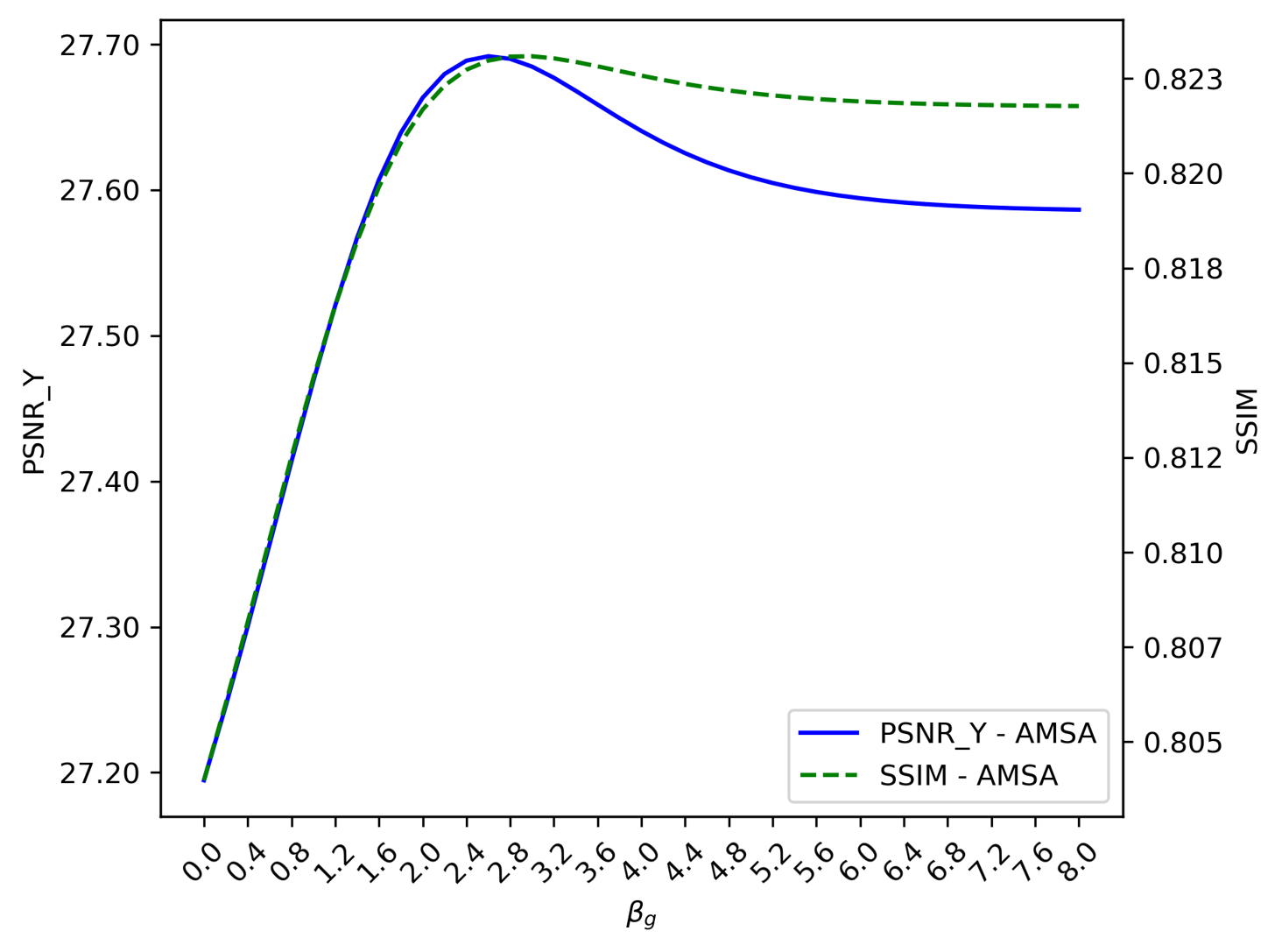}
    \caption{}
    \end{subfigure}
    \begin{subfigure}[b]{0.31\columnwidth}
    \centering
    \includegraphics[width=\textwidth]{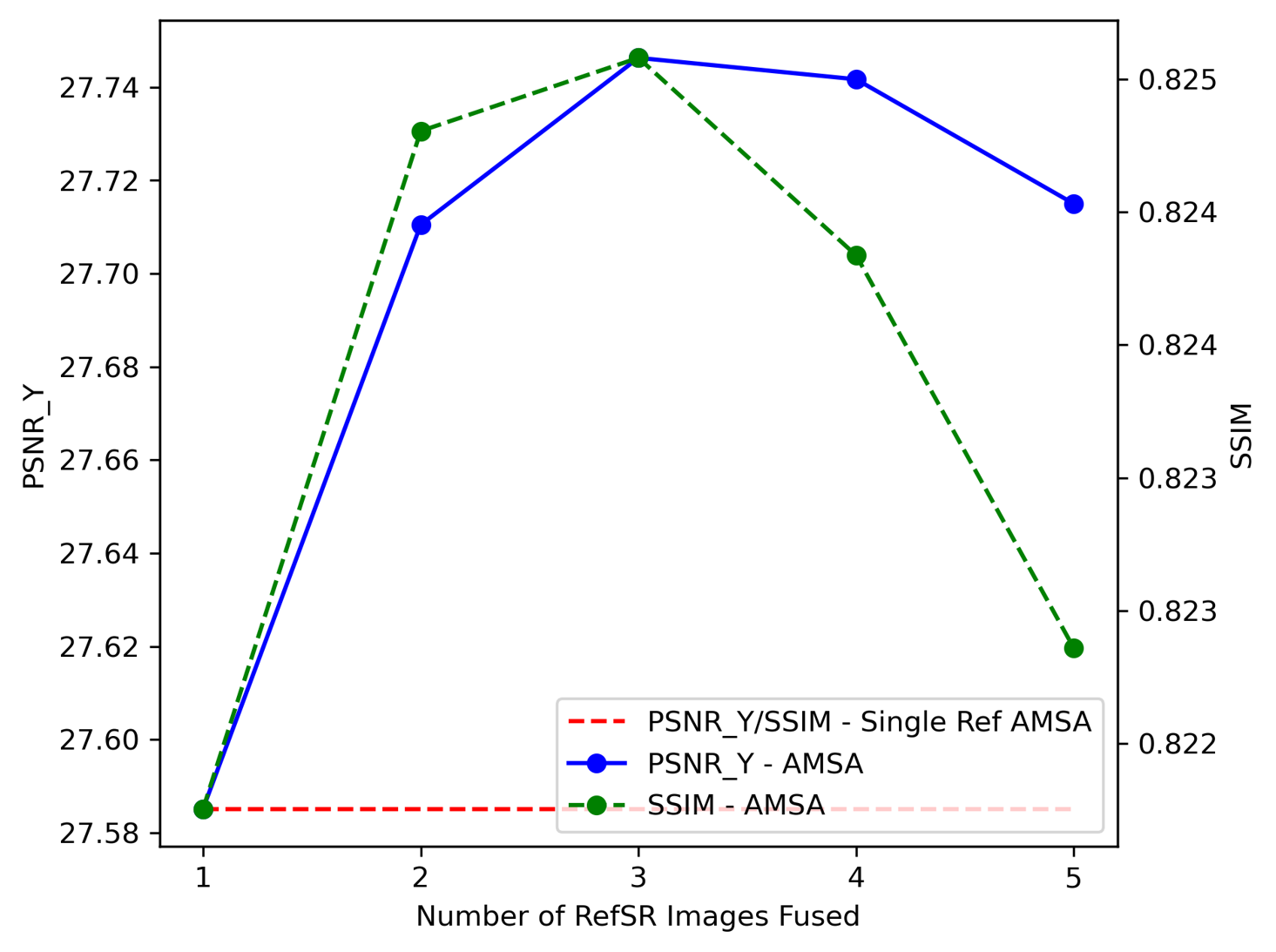}
    \caption{}
    \end{subfigure}
    
    \caption{Quantitative evaluation by applying the proposed method to C$^2$-Matching and AMSA. (a) \& (d) Performance changes w.r.t. $\beta$. (b) \& (e) Performance changes w.r.t. $\beta_g$. (c) \& (f) Performance changes w.r.t. number of images fused.}
    \label{fig:metrics}
\end{figure}

\begin{figure}[ht]
    \centering
    \begin{subfigure}[b]{0.46\columnwidth}
    \centering
    \includegraphics[width=\textwidth]{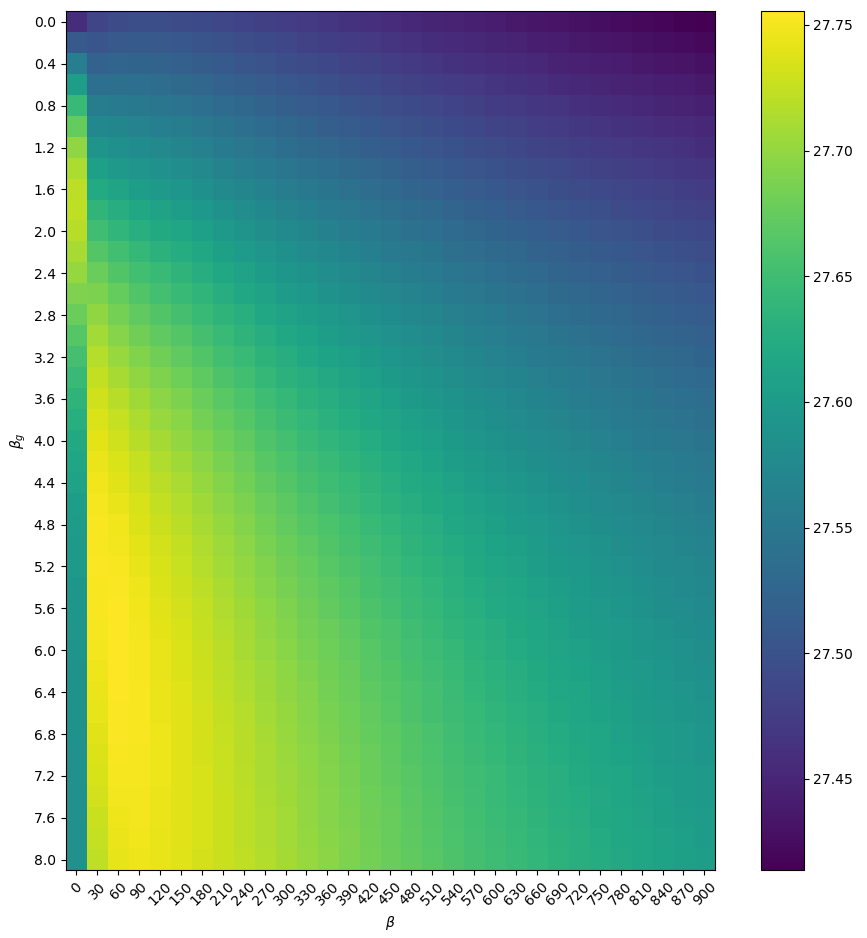}
    \end{subfigure}
    \begin{subfigure}[b]{0.46\columnwidth}
    \centering
    \includegraphics[width=\textwidth]{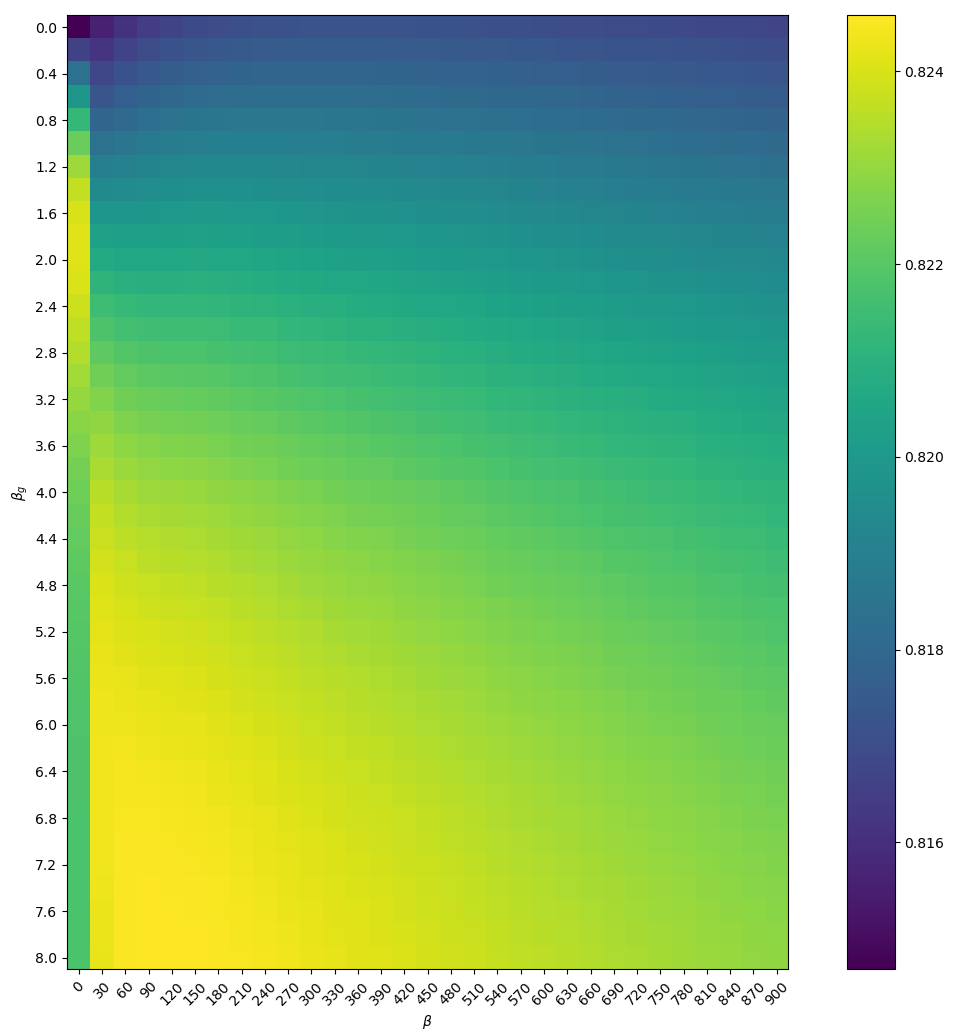}
    \end{subfigure}
    
    \caption{The PSNR\_Y (left) and SSIM (right) heatmap with varied parameters $\beta$ and $\beta_g$}
    \label{fig:heatmap_beta_beta2}
\end{figure}

\noindent \textbf{Evaluate Adaptive Weight Masking.} We applied the proposed method to $C^2$-Matching and AMSA to perform quantitative evaluations. To analyze the effectiveness of Adaptive Weight Masking, we fixed $\beta_g$ to be $0$, which essentially disables the effect of the Global Reference-Quality-based weight.  Fig. \ref{fig:metrics} (a) and (d) shows how the PSNR\_Y/SSIM changes as $\beta$ varies from $0$ to $810$. Note that when $\beta = 0$, the method degrades to Naive Fusion. It can be observed that both PSNR\_Y and SSIM get better as $\beta$ increase, that is, as the distorted pixels are penalized more and more. While PSNR\_Y begins to decrease as $\beta$ passes the optimal value, SSIM keeps increasing and then decreases mildly compared with PSNR\_Y. While the increased PSNR\_Y/SSIM validates the effectiveness of Adaptive Weight Masking, it should be noticed that the PSNR\_Y/SSIM is worse than the $27.16/0.805$ (see Table \ref{tab:quanti}) achieved by $C^2$-Matching and $xx$ achieved by AMSA using the single most relevant reference image. That is due to the Adaptive Weight Masking's insensitiveness to relatively small distortions in SR results, as stated in Sec. \ref{sec:global_weight}.
\\
\noindent \textbf{Evaluate Global Reference-Quality-based Weight.} As shown in Fig. \ref{fig:metrics} (b) and (e), as $\beta_g$ increases from 0 to 8 the PSNR\_Y/SSIM of both $C^2$-Matching and AMSA first arise then plateau, given the condition that $\beta$ is fixed to 0. When $\beta_g = 0$ the process is equivalent to Naive Fusion while $\beta_g \to \infty$ would be the case of using the single best SRefSR result alone. This trend shows that there is indeed additional valuable information from the sub-optimal SRefSR results, and even a simple trick as weighted averaging would improve the image quality with a properly chosen $\beta_g$.
\\
\noindent \textbf{Evaluate the Combined Weighting Method} Fig. \ref{fig:heatmap_beta_beta2} how the metrics values changes with varied $\beta$ and $\beta_g$, and only the results for ASMA are demonstrated since the evaluation with C$^2$-Matching produce similar results. It can be observed there is a sharp change when $\beta$ changes from $0$ to $30$, which is consistent with the previous observation on the quick increase in metrics values in Fig. \ref{fig:metrics} (a) and (d).
\\
\noindent \textbf{Evaluate the Number of Images Fused.} In Fig. \ref{fig:metrics} (c) and (f), we cumulatively fuse more and more RefSR images to and show the resulting image's PSNR\_Y/SSIM. Notice that the order of fusion is chosen such that the best-quality RefSR image is taken to be fused with RefSR images that are of less and less quality. It can be shown that our proposed method is resistant to imperfect RefSR images since the PSNR\_Y and SSIM starts to decrease only after the fourth RefSR image is fused and PSNR\_Y decrease mildly.
\\
It is noteworthy that the evaluation above is done separately with two different state-of-the-art SRefSR models, namely $C^2$-Matching and AMSA, and a consistent behavior is observed. Therefore we claim that the proposed method has the potential to be incorporated into a variety of SRefSR pipelines to get a performance improvement.

\subsection{Case Study with SelfDLSR}
The models in the previous experiment use HR references to generate super-resolution images, so we have also studied the effectiveness of this super-resolution pipeline with other types of references. In particular, we experimented with SelfDZSR which uses telephotos (zoom-in images) as references.

\noindent \textbf{Dataset Preparation.} While the CUFED5 test set can be directly adopted for AMSR and $C^2$-matching, it does not contain telephotos that are required for the DZSR model. Therefore, an image processing pipeline is constructed to generate short-focus and telephotos from CUFED5. This takes the ground truth high-resolution images alone and outputs the short-focus low-resolution images by 4x bicubic downsampling, the same as the previous experiment with $C^2$-Matching and AMSA. It also renders telephotos by cropping the high-resolution images to simulate the effect of zooming in. This pipeline is flexible because it can support any image dataset and different resize factors for more comprehensive comparisons.

\begin{figure}[ht]
\centering
\includegraphics[width=0.9\columnwidth]{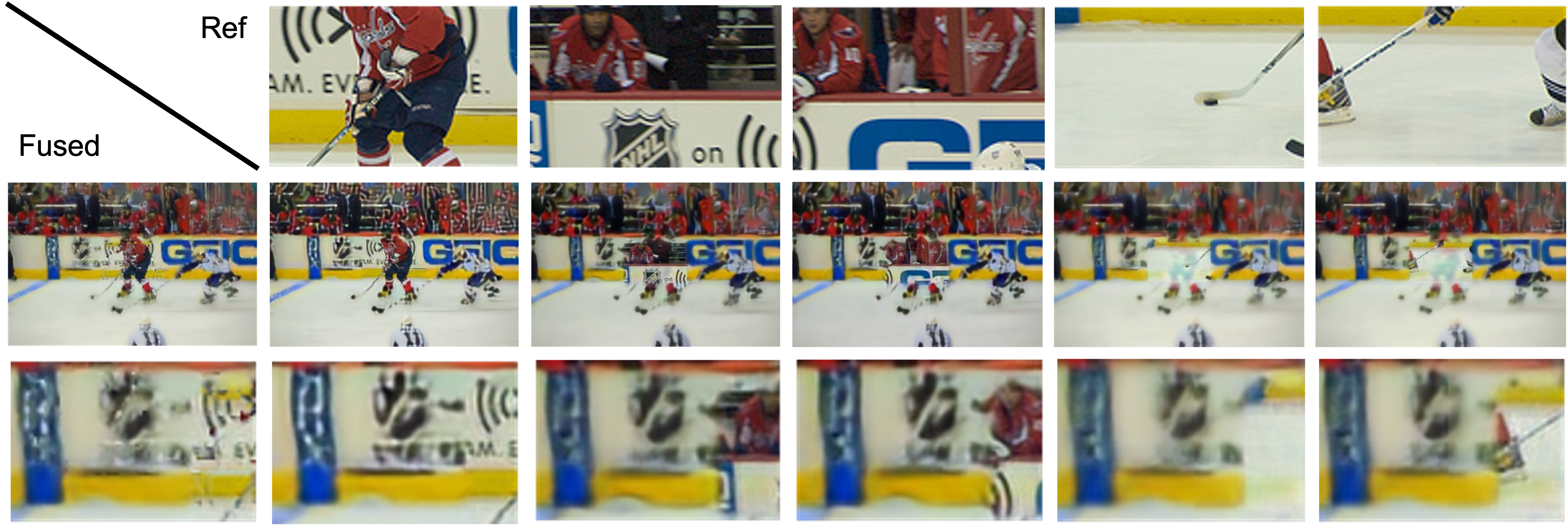}
\caption{Qualitative Comparisons by using different reference images (telephoto) for SelfDZSR. The leftmost one is the fused image while the right 5 are outputs from individual references.}
\label{fig:SelfDZSRoutput}
\end{figure}

\noindent \textbf{Qualitative Comparisons.} Figure \ref{fig:SelfDZSRoutput} shows the qualitative results. When center reference (telephoto) are used, we can see that the resulting images show obvious artifacts in the surrounding of the output images. However, when we use the telephotos taken from the corners of short-focused images, the center part of the super-resolution images seems to be replaced by the reference, largely deviating from the ground truth. This is different from what we expected as the reference patches used during training are not restricted to the center but at randomized positions with simple augmentation (flipping and $90 \degree$ rotation), making it resilient to the displacement of reference. Note that the surrounding parts are smoother without the artifacts. Nonetheless, the fused output (the leftmost one) can combine the smoothness from non-center references and the overall structure with center references, showing significant improvement from each of the single-reference super-resolution outputs. 

\begin{figure}[ht]
\centering
\includegraphics[width=0.9\columnwidth]{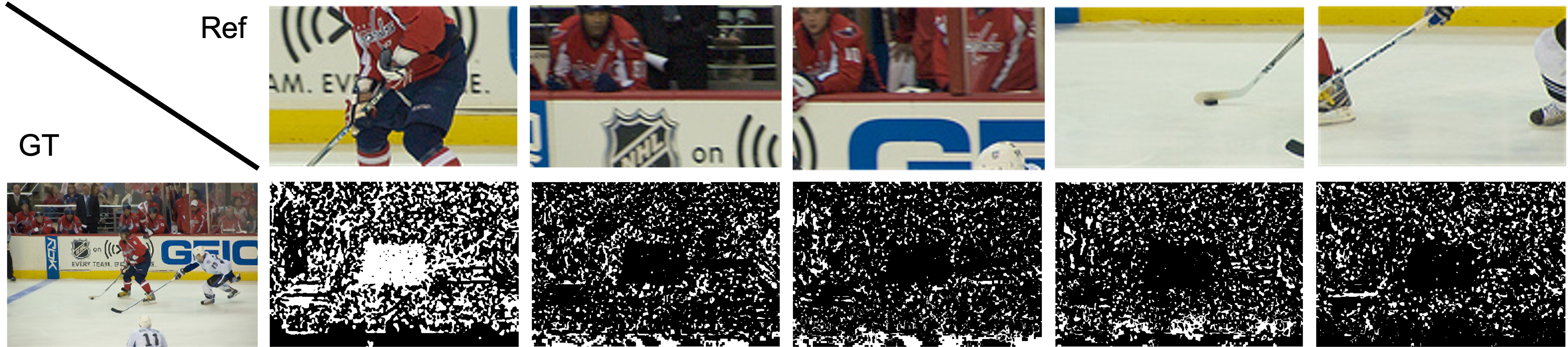}
\caption{Binary weight masks from Global Reference-Quality-based Weight by using different reference images (telephoto) for SelfDZSR.}
\label{fig:SelfDZSRmask}
\end{figure}

\noindent \textbf{Weight Analysis.} 
Figure \ref{fig:SelfDZSRmask} illustrates the binary weight masks for different references respectively, we can see that in the first reference (using a center telephoto), most of the pixels in the center part have the highest weights across references, showing that the Adaptive Weight Masking step in the pipeline can identify the distortion in center parts and put lower weights. We also discover that some of the pixels in the surrounding area are the highest in non-center reference (the second to fifth in the figure), showing that the step can recognize the artifacts in the surroundings. Therefore, the fusion module can indeed combine the smoothness from non-center references and the overall structure with center references.

\section{Discussion}
To improve our image fusion module, we will try to explore better ways to assign pixel and image weights for Adaptive Weight Masking and Global Reference-Quality-base. In particular, we will adopt other priors that combine the sharp region of each image to provide a more fine-grained result. This can be gradient-based methods such as Poisson image processing and pyramid-based methods such as Gaussian pyramid.

We can also adopt neural network models for image fusion such that the prior knowledge of what a "natural image" is learned by the neural network to select the best-reconstructed regions in each RefSR image. This approach has the potential to significantly outperform our simple global reference-quality-based weighting strategy.

\bibliographystyle{IEEEtran}
\bibliography{references}

\end{document}